\documentclass[preprint,review,12pt]{elsarticle}

\usepackage{amssymb}
\usepackage{amsmath}
\usepackage{multirow}
\usepackage{booktabs}

\usepackage{xcolor}
\usepackage{nicematrix}
\usepackage{ulem}

\newcommand{\rev}[1]{\textcolor{black}{#1}}

\journal{Pattern Recognition}

\begin{document}

\begin{frontmatter}


\title{Domain-generalizable Face Anti-Spoofing with Patch-based Multi-tasking and Artifact Pattern Conversion}
\author[1,2]{Seungjin Jung\fnref{fn1,fn2}}
\author[2]{Yonghyun Jeong\fnref{fn1}}
\author[2,3]{Minha Kim\fnref{fn2}}
\author[2,4]{Jimin Min\fnref{fn2}}
\author[1,2]{Youngjoon Yoo}
\author[1,5]{Jongwon Choi\corref{cor1}}
\ead{choijw@cau.ac.kr}

\affiliation[1]{organization={Department of Artificial Intelligence},
            addressline={Chung-Ang University}, 
            city={Seoul},
            postcode={06974}, 
            country={Korea}}

\affiliation[2]{organization={Team of Image Vision},
            addressline={Naver Cloud}, 
            city={Seongnam},
            postcode={13561}, 
            country={Korea}}

\affiliation[3]{organization={Department of Human-centric AI},
            addressline={Nota AI}, 
            city={Seoul},
            postcode={06164}, 
            country={Korea}}
            
\affiliation[4]{organization={Department of Computer Engineering},
            addressline={Hanbat National University}, 
            city={Daejeon},
            postcode={34158}, 
            country={Korea}}
            
\affiliation[5]{organization={Department of Imaging Science, GSAIM},
            addressline={Chung-Ang University}, 
            city={Seoul},
            postcode={06974}, 
            country={Korea}}
            
\cortext[cor1]{Corresponding author}
\fntext[fn1]{Seungjin Jung and Yonghyun Jeong g contributed equally to this work.}
\fntext[fn2]{Authors Seungjin Jung, Jimin Min, and Minha Kim conducted this research during their internship at Naver Cloud.}
\tnotetext[1]{© 2026. This manuscript version is available under the CC BY-NC-ND 4.0 license. The published version is available at DOI: https://doi.org/10.1016/j.patcog.2026.113640}

\begin{abstract}
Face Anti-Spoofing (FAS) algorithms, designed to secure face recognition systems against spoofing, struggle with limited dataset diversity, impairing their ability to handle unseen visual domains and spoofing methods. We introduce the Pattern Conversion Generative Adversarial Network (PCGAN) to enhance domain generalization in FAS. PCGAN effectively disentangles latent vectors for spoof artifacts and facial features, allowing to generate images with diverse artifacts. We further incorporate patch-based and multi-task learning to tackle partial attacks and overfitting issues to facial features. Our extensive experiments validate PCGAN's effectiveness in domain generalization and detecting partial attacks, giving a substantial improvement in facial recognition security.
\end{abstract}



\begin{keyword}
Domain Generalizable Face Anti-Spoofing\sep 
Pattern Conversion GANs\sep 
Disentangle Texture and Contents\sep
Patch based Multi Task Learning
\end{keyword}

\end{frontmatter}

\section{Introduction}
\label{introduction}
Face recognition technology has made remarkable progress in recent years and is now widely deployed in various security systems.
However, these recognition systems remain vulnerable to \rev{spoofing} attacks in which adversaries deceive the system using \rev{either direct or indirect manipulation strategies~\cite{jiang2025learning, antil2025unmasking}.
Direct attacks, also known as presentation attacks, target the camera sensor by presenting physical artifacts such as printed photos or replayed videos. In contrast, indirect attacks manipulate data after acquisition, including digital attacks such as deepfake-based identity manipulation. In practice, attackers typically lack access beyond the input acquisition stage~\cite{jiang2025learning}.}
\rev{Therefore}, Face Anti-Spoofing (FAS) algorithms have been developed to \rev{detect presentation attacks by distinguishing live face from spoofed face presented through physical media, such as printed photographs or electronic displays.} 

Although early FAS models have shown effectiveness in controlled scenarios, their performance degrades when exposed to unseen capturing environments ({\it{e.g.} brand new camera, irregular lighting conditions}) and new types of presentation attacks. 
The challenge lies in the lack of diversity in both subject identities and capture environments in existing training datasets, which hinders the generalization of FAS algorithms to real-world scenarios.
Commonly used FAS datasets~\cite{zhang2012face,Chingovska_BIOSIG-2012,Diwen2014,OULU_NPU_2017} typically contain fewer than 100 identities and lack diversity in capture environments, such as types of recapturing devices and illumination conditions.

Since it is infeasible to cover all possible scenarios involving diverse capturing environments and presentation attacks, researchers have improved the generality of detectors based on two major approaches: domain adaptation and domain generalization.
Domain Adaptation for FAS (DAFAS)~\cite{guo2022multi,zhou2022generative} incorporates domain adaptation techniques into face anti-spoofing by leveraging data from a specific source domain to progressively adapt the model to a distinct target domain.
In contrast, Domain Generalizable FAS (DGFAS)~\cite{sun2023rethinking,srivatsan2023flip,ma2024dual,huang2023face} trains a FAS model to learn domain-invariant features using a multi-source domain dataset without relying on target domain data during training.

DAFAS achieves strong performance despite the absence of target-domain labels; however, it requires a dual-phase training procedure and access to target-domain data, which is often impractical in real-world scenarios. In contrast, DGFAS learns more generalized liveness and spoof features by leveraging multiple training domains, enabling competitive performance without target-domain data or additional model adaptation. Nevertheless, the limited diversity of existing FAS datasets still restricts generalization in unconstrained environments.

To address this limitation, we propose the Pattern Conversion Generative Adversarial Network (PCGAN), which disentangles spoof artifacts and facial content to generate diverse synthetic FAS images with different artifact types. This strategy enriches training data diversity and enhances domain-generalizable feature learning. In addition, we incorporate patch-based learning and multi-task learning in the detector to handle partial spoofing and mitigate identity overfitting. Patch-based learning adopts a self-supervised scheme with randomly sampled artifact regions, while multi-task learning jointly exploits full-face images and synthesized patches during training.

The key contributions of our work are given as follows\footnote{The source code and dataset will be available upon publication.}:
\begin{itemize} \label{contribution}
\item We propose Pattern Conversion Generative Adversarial Network (PCGAN), which effectively disentangles latent vectors corresponding to spoof artifacts and facial contents.
\item By using disentangled features, we generate images that combine various spoof artifacts with a single facial identity, thereby enhancing the diversity of training data for Face Anti-spoofing (FAS).
\item We propose patch-based and multi task learning methods to overcome partial spoofing attacks and the overfitting issue to well-aligned and limited face images.
\item Through extensive experiments, we verify the effectiveness of our method not only for domain generalization but also for partial attack detection.
\end{itemize}

\section{Related work}
In this chapter, we discuss previous studies on DGFAS tasks, including data augmentation techniques using generative models for FAS and the patch-based FAS training approach. We present a comparison of our approach with these prior works.

\subsection{Domain Generalization for FAS}
The field of Face Anti-Spoofing (FAS) has seen several notable studies addressing the challenges associated with dataset diversity and domain generalization. In this section, we discuss relevant works that have contributed to the advancement of FAS techniques.
Huang~\cite{huang2023face} introduced global attention learning, while Wang~\cite{wang2022face} and Liao~\cite{liao2023domain} proposed transformer-based learning methods for domain invariance. 
\cite{wang2023consistency} utilized consistency regularization to perform domain generalization. 
They separated attack samples by domain and learned decision boundaries, while aggregating live samples across domains, resulting in a more generalizable feature space.
\rev{Despite these advances, luminance and lightness variations across capture environments still cause substantial shifts in feature distributions, leading to inconsistent representations of spoof samples. 
To address this issue, we adopt an image-level approach that combines a generative model for learning spoofing artifacts with a patch-based learning strategy focusing on localized image regions.}

\subsection{Generative model for FAS}
The field of Face Anti-Spoofing (FAS) faces significant challenges due to limited dataset diversity and the need for robust domain generalization. Although several FAS datasets have been released~\cite{zhang2012face,Chingovska_BIOSIG-2012,Diwen2014,OULU_NPU_2017}, they do not cover the wide range of spoofing techniques used by attackers. Moreover, the rapid advancement of display technology further complicates the FAS landscape. Unlike face recognition, which utilizes large-scale datasets with thousands of identities, FAS datasets typically include fewer than 100 identities, resulting in models that struggle to generalize to new, unseen data.
To address these issues, generative models have been employed to overcome dataset diversity limitations. Techniques such as StyleGAN~\cite{menon2019style}, domain adaptation~\cite{nikisins2019domain}, and StyleAssemble~\cite{styleassemble} have been proposed to enhance generalization by generating data with domain-specific information. For style transfer, \cite{yadav2021cit} used a cycle consistency approach, while \cite{styleassemble} employed Adaptive Instance Normalization (AdaIN).
\rev{Beyond style transfer, recent studies have explored generative modeling for FAS, where \cite{long2024generalized} learn intrinsic liveness characteristics via real-face generation and \cite{antil2024securing} leverage lightness-aware representations for image synthesis. In contrast, we propose a Pattern Conversion GAN that converts spoof artifact patterns between live and spoof images.
While prior generative approaches mainly focus on domain style adaptation or one-directional artifact removal, our network explicitly disentangles and manipulates spoof artifacts. By separating facial content from artifact patterns and recombining them across samples, our method increases spoof data diversity at the artifact level, thereby improving robustness and generalization.}

\subsection{Patch-based Learning for FAS} 
Face recognition tasks exhibit high accuracy, but face images are susceptible to various attack types. To address these vulnerabilities, \cite{atoum2017face,9484395,9796574,Shen_2019_CVPR_Workshops} proposed a FAS method using local features and depth maps of face images. However, their approach involved extracting small patches from face images and training them separately with a depth-based CNN model, resulting in no parameter sharing between models. \cite{chuang2023generalized} overcame this limitation with a multi-task meta-learning framework using a U-net-based face parsing module and depth estimator for spoof classification, facilitating parameter sharing.

Studies by PatchNet~\cite{patchnet,yu2024rethinking}, and \cite{yang2020pipenet} demonstrated the benefits of incorporating patch-level inputs in training data. \cite{srivatsan2023flip} achieved success by employing CLIP\cite{radford2021learning}, enhancing data diversity and enabling models to learn features specific to local regions targeted in spoof attacks.
Another recent work utilizing a multi-modal approach, CFPL-FAS \cite{liu2024cfpl}, addresses the problem of domain generalization for FAS through textual prompt learning conditioned on content and style features.%
Inspired by these studies, our paper introduces auxiliary supervision by training a FAS model using patches based on the CLIP model, recognizing that relevant information can be derived from both face images and external parts through patch-based learning.

\section{Methods\label{sec.3}}
In this chapter, we introduce two networks to perform domain generalization in Face Anti-Spoofing (FAS) tasks.
The first network is Pattern Conversion Generative Adversarial Networks (PCGAN), which can convert existing artifacts in the attack image to the target image or remove existing artifacts. The second network is a Patch-based Multi-tasking Network (PMN) to robustly detect unseen presentation attacks.

\begin{figure*}[!t] 
\centering
\includegraphics[width=0.95\linewidth]{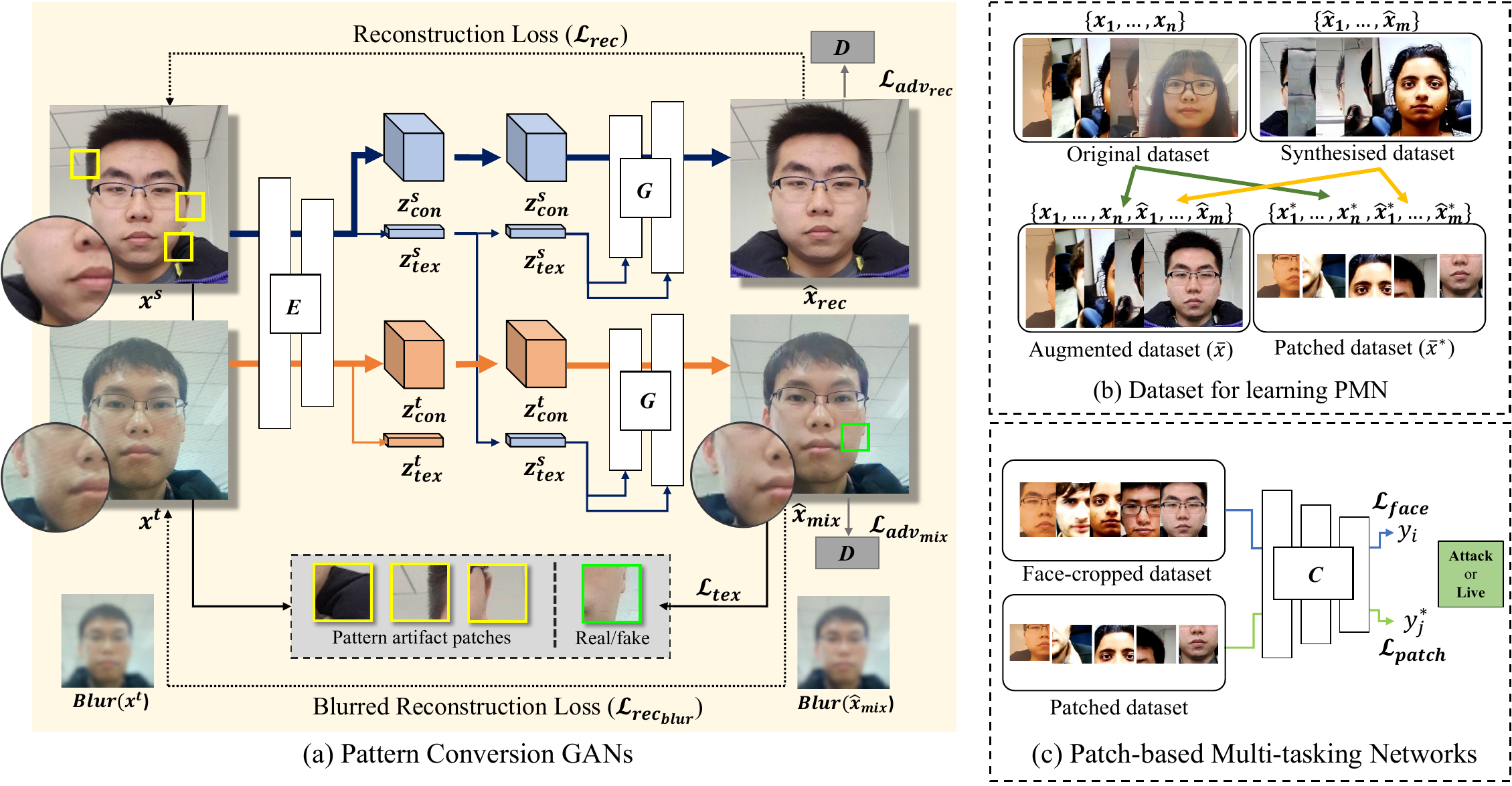}
\vspace{-5mm}
\caption{Overall framework. (a) shows the disentanglement and conversion of artifact patterns \rev{by combining spatial content features and artifact representations from different images.}
\rev{(b) PCGAN-based data augmentation and construction of patched datasets for training. (c) Patch-based Multi-tasking Network (PMN), which jointly learns from patched and full face-cropped images.}
}
\label{fig:framework}
\vspace{-3mm}
\end{figure*}

\subsection{Pattern Conversion GANs}\label{sec:patchgan}
The PCGAN aims, in the context of presentation attacks, to disentangle the presentation artifacts from the image elements excluding the artifacts. Then, in the following step, we could transfer these disentangled artifacts onto live images or remove the artifacts from presentation images.

\noindent\textbf{Spoofing Artifact:} \rev{
Spoofing artifacts are commonly understood as visual patterns introduced by the attack medium and the re-capturing process, rather than by genuine facial characteristics.
Specifically, print attacks often exhibit printing-process artifacts such as halftone dot patterns~\cite{cai2025towards}, while replay attacks may produce moiré-like interference caused by the interaction between display pixel structures and camera sensor sampling grids~\cite{wang2025fsfm}.
}

\subsubsection{Architecture}
\rev{
In this study, we propose a model that explicitly extracts spoofing artifacts from presentation attack images and exploits them in two complementary ways: removing artifacts from attack images and transferring them onto real images. Since spoofing artifacts do not naturally appear in genuine facial images, contrasting real and attack images enables reliable separation of artifact patterns from facial content.
Based on this observation, our framework builds upon a swapping auto-encoder architecture~\cite{SAE}, with a modified encoder that employs reduced downsampling to preserve fine-grained spoofing artifacts. The proposed model comprises four main components: an encoder, a generator, a discriminator, and a patch discriminator.}
The overall framework is illustrated in Fig.~\ref{fig:framework}-(a).

\noindent\textbf{Encoder:}
\rev{
The encoder $E$ maps an input image $x \in \mathbb{R}^{3 \times 1024 \times 1024}$ into two disentangled latent representations: an artifact pattern representation $z_{pat} \in \mathbb{R}^{8 \times 512 \times 512}$ and a facial content representation $z_{con} \in \mathbb{R}^{8}$. 
Unlike conventional encoders that employ multiple downsampling layers, our encoder applies downsampling only once to better preserve fine-grained spoofing artifact patterns. Specifically, we follow the architecture of~\cite{SAE} but reduce the parameter $netE\_num\_downsampling\_sp$ from 4 to 1, as excessive downsampling significantly degrades spatial resolution and leads to the loss of critical artifact information.
}

\noindent\textbf{Generator:}
The $z_{\rev{pat}}$ and $z_{con}$ of different dimensions derived from the encoder are used as inputs to the generator. We feed-forward the content that contains spatial information to the generator to effectively reconstruct the input image. On the other hand, the latent vectors for artifact patterns are given to the generator as the input to every CNN block through adaptive instance normalization
for the integration of two disentangled latent vectors~\cite{stylegan}.

\noindent\textbf{Discriminator:}
In order to maximize the fidelity of the image generated by the generator, we use the discriminator proposed by \cite{stylegan}. We also include an additional network called patch discriminator to construct an artifact-aware pattern conversion model.

\subsubsection{PCGAN Losses}
\noindent\textbf{Pattern Conversion Loss:}
Pattern conversion loss is designed to disturb the recognition between fake converted artifact patterns and original artifact patterns when artifact patterns in an image are converted. The pattern patch discriminator is trained not to distinguish between the artifacts in the mixed $\hat{x}^{mix}$ determined by $z^{\rev{src}}_{\rev{pat}}$ encoded from $x^{\rev{src}}$. Therefore, the pattern conversion loss is defined as follows:

\begin{equation}
    \begin{aligned}
        &\mathcal{L}_{\rev{pat}}(E,G,D_{patch})=\\
        &\mathop{\mathbb{E}}\limits_{\substack{x^{\rev{src}}, x^{\rev{tgt}}\sim\mathbf{X}, x^{\rev{src}}\neq x^{\rev{tgt}}}}\left[-\log\left(D_{patch}(\text{crop}(G(z^{\rev{tgt}}_{con},z^{\rev{src}}_{\rev{pat}})), \text{crop}(x^{\rev{src}}))\right)\right],
    \end{aligned}
\label{eq:ce_patch}
\end{equation}

where $\text{crop}()$ function defines the operation of the patch-based crop from the given image, and $(z^{\rev{tgt}}_{con}, z^{\rev{tgt}}_{\rev{pat}})=E(x^{\rev{tgt}})$ and $(z^{\rev{src}}_{con}, z^{\rev{src}}_{\rev{pat}})=E(x^{\rev{src}})$.

\noindent\textbf{Reconstruction Loss:}
We construct a reconstruction loss to ensure that the network preserves all the information of the input image. The image reconstruction loss, which allows the network to reproduce the same image from the input image $x\sim\textbf{X}\subset\mathbb{R}^{224\times 224\times 3}$, is expressed as follows.
\begin{equation}\label{eq:rec_loss}
        \mathcal{L}_{rec}=\mathbb{E}_{x\sim \mathbf{X}}\left[||x-G(E(x))||_{2}\right].
\end{equation}

\rev{
Conventional swapping auto-encoders preserve the overall facial structure but often introduce unintended variations in fine-grained components such as the eyes, nose, and mouth. To improve robustness to such local structural changes, we introduce a blurred reconstruction loss that preserves the semantic content of the mixed image $x^{mix}$.
Specifically, we apply a downsampling-based blurring operation that suppresses spoofing artifact patterns while retaining semantic facial information. The input resolution is reduced from $1024 \times 1024$ to $512 \times 512$, preserving global facial structure while effectively removing high-frequency artifact components. Since only a $1/2$ downsampling is applied, major facial structures remain largely intact.
Based on this property, we enforce that images sharing the same content representation $z_{con}$ but different artifact representations $z_{pat}$ converge to the same representation in the blurred space. The blurred reconstruction loss is defined as follows:}
\begin{equation}\label{eq:rec_blur_loss}
\begin{aligned}
\mathcal{L}_{rec_{blur}}=\mathbb{E}_{\substack{x^{\rev{src}}, x^{\rev{tgt}}\sim \mathbf{X},\\ x^{\rev{src}}\neq x^{\rev{tgt}}}}\left[||\text{blur}(x^{\rev{tgt}})-\text{blur}(G(z^{\rev{src}}_{\rev{pat}},z^{\rev{tgt}}_{con}))||_{2}\right],
\end{aligned}
\end{equation}
\rev{where blur($\cdot$) denotes a blurring operation implemented via down sampling.}


\noindent\textbf{Adversarial Loss:}
The purpose of the adversarial loss is to maintain the visual fidelity of the reconstructed images $\hat{x}^{rec}$ and mixed images $\hat{x}^{mix}$, as shown below:
\begin{equation}\label{eq:adv_mix_loss,eq:adv_rec_loss}
\begin{aligned}
\mathcal{L}_{adv_{rec}}(E,G,D) &=\mathop{\mathbb{E}}\limits_{x\sim \mathbf{X}}\left[-\log(D(G(E(x^{\rev{src}}))))\right],\\
\mathcal{L}_{adv_{mix}}(E,G,D) &=\mathop{\mathbb{E}}\limits_{\substack{x^{\rev{src}}, x^{\rev{tgt}}\sim \mathbf{X},\\ x^{\rev{src}}\neq x^{\rev{tgt}}}}\left[-\log(D(G(E(x^{\rev{src}},x^{\rev{tgt}}))))\right]. 
\end{aligned}
\normalsize
\end{equation}

This resolves the incongruity of the mixed image and generates a realistic image.

\noindent\textbf{Total Loss:}
In each iteration of the training phase, the model conducts mixture and reconstruction, simultaneously.
The total loss \rev{is obtained by combining Eqs.~(1), (2), (3), and (4), and is} formulated as follows:
\begin{equation}\label{eq:total_loss}
\begin{aligned}
        \mathcal{L}_{PCGAN}=\mathcal{L}_{rec}+\mathcal{L}_{rec_{blur}}+\mathcal{L}_{adv_{rec}}+\mathcal{L}_{adv_{mix}}+\mathcal{L}_{\rev{pat}}.
\end{aligned}
\normalsize
\end{equation}

\subsubsection{Artifact Pattern Conversion}
\rev{To generate training samples for PMN, we augment the data using ground-truth labels. Given attack images $\{x^{a}\}_{i=1}^{n}$, a synthetic attack image $\hat{x}^{\text{atk}}$ is produced by transferring spoofing artifact patterns from an attack image $x^{a}_{h}$ onto a live image $x^{\text{liv}}_{g}$, where $g,h \in [1,\ldots,n]$. 
Conversely, a synthetic live image $\hat{x}^{\text{liv}}$ is obtained by removing spoofing artifacts from an attack image and replacing them with live facial content from $x^{\text{liv}}$. 
This symmetric conversion enables balanced augmentation for both classes while preserving the original class ratio.}

\subsection{Patch-based Multi-tasking Network}
Patch-based Multi-tasking Network (PMN) consists of CLIP~\cite{radford2021learning} $C$, Multi Layer Perceptron (MLP) $F$, and Fully Connected Layer (FC) $M$, and we train PMN by the label $y \sim \textbf{Y}\subset\mathbb{R}^2$, the description text $t\sim \textbf{T}$, and the combined image $\bar{x}\sim\bar{\textbf{X}}$ which is a combination of the generated image $\hat{x}\sim\hat{\textbf{X}}$ and the original image $x\sim\textbf{X}$. \rev{All images are resized to 224×224.}

\subsubsection{Overall Framework}
The use of patch-based models, as demonstrated in previous studies~\cite{
atoum2017face,patchnet
}, can increase the diversity of data and encourage the network to learn features specific to presentation attacks in localized areas. 
We use patched images $\bar{x}^*\sim\bar{\textbf{X}}^*$ obtained by randomly cropping images including parts of the face from the whole image $\bar{x}$. The methods for constructing patched images and augmented images for PMN training are shown in Fig~\ref{fig:framework}-(b).

Multi-Task Learning (MTL) aims to share knowledge while simultaneously learning data with similar but different tasks.
We employ the CLIP~\cite{radford2021learning} with an MLP to share knowledge in MTL, learning $\bar{x}^*$ and $\bar{x}$ simultaneously.
Therefore, PMN not only learns presentation attack characteristics in full-face images but also learns spatial consistency between background and foreground regions. The last fully connected layer of PMN separates each of the two tasks. After the training, only the face-cropped image is used during inference.
The overall framework is shown in Fig~\ref{fig:framework}-(c).

\begin{table*}[t]
\caption{\label{tab:prompt}\rev{\textbf{Text Prompt Templates for Live and Attack Faces.}}}
\centering
\resizebox{0.8\linewidth}{!}{
\begin{NiceTabular}{@{}c|c|c@{}}
\hline
Prompt No. & Liveness Face Prompts & Attack Face Prompts \\
\hline
No.1 & This is an example of a real face & This is an example of a spoof face \\
No.2 & This is a bonafide face & This is an example of an attack face \\
No.3 & This is a real face  & This is not a real face \\
No.4 & This is how a real face looks like & This is how a spoof face looks like \\
No.5 & a photo of a real face & a photo of a spoof face \\
No.6 & This is not a spoof face & a printout shown to be a spoof face \\
\hline
\end{NiceTabular}}
\vspace{-3mm}
\end{table*}

\subsubsection{PMN Losses}
\noindent\textbf{CLIP Loss:}
Since previous work~\cite{srivatsan2023flip} shows that CLIP~\cite{radford2021learning} is effective for the Presentation Attack Detection task, we use CLIP as the backbone network. 
CLIP is trained on a large set of image and text pairs taken from the Internet and is fine-tuned to suit the task when used.
CLIP consists of image encoder $C_I$ and text encoder $C_T$, where description texts $t\sim T$ for each label similar to~\cite{srivatsan2023flip}.
The description texts consist of six sentences~\rev{\cite{srivatsan2023flip}} each for attack and live \rev{depicted as in Tab.~\ref{tab:prompt}}, and they are transformed into text embedding features through a text encoder $C_T$.
We calculate the average for each attack and live feature embedding and the cosine similarity between the mean text and image embedding features.
Thus, we can design CLIP loss as follows:
\begin{equation}\label{eq:clip}
\begin{aligned}
        \mathcal{L}_{clip} = \mathop{\mathbb{E}}\limits_{(\bar{x},t,y)\sim{\bar{\mathbf{X}},\mathbf{T},\mathbf{Y}}}
        \Big[CE\big(\langle C_I(\bar{x}), \mathop{\mathbb{E}}\limits_{t_{a},t_{l}\sim T}[C_T(t)]\rangle,y\big)\Big],
        \end{aligned}
\normalsize
\end{equation}
where $CE$ is a cross-entropy loss, $\langle,\rangle$ represents an inner product, and $t_{a}$ and $t_{l}$ mean text description elements for attack and live, respectively.

\noindent\textbf{Multi-tasking Loss:}
During the training of PMN, we utilize both the full image, denoted as $x$, and its corresponding patch image that is $x^*$ and obtained by cropping from the full image $x$ as a random crop size scale $0.2$-$1.0$. 
\rev{Patch-level supervision using randomly cropped images $x^*$ enhances robustness to partial attacks by enabling localized spoofing artifacts to directly contribute to the detection decision.}
Then, we apply the MLP approach to extract image features from the embedded image feature of CLIP~\cite{dosovitskiy2020image,srivatsan2023flip}.
Thus, we formulate PMN Loss defined as:
\begin{equation}\label{eq:ce_patch}
\begin{aligned}
        \mathcal{L}_{face} = \mathop{\mathbb{E}}\limits_{(\bar{x},y)\sim (\bar{\mathbf{X}},\mathbf{Y})}
        \Big[CE\big(M_1(F(C(\bar{x}))),y\big)\Big],
        \end{aligned}
\end{equation}
\begin{equation}\label{eq:ce_whole}
\begin{aligned}
        \mathcal{L}_{patch} = \mathop{\mathbb{E}}\limits_{(\bar{x}^*,y)\sim (\bar{\mathbf{X}}^*,\mathbf{Y})}
        \Big[CE\big(M_2(F(C(\bar{x}^*))),y\big)\Big].
\end{aligned}
\end{equation}

\noindent\textbf{Center Loss:}
Center loss~\cite{wen2016discriminative} 
is adopted to optimize the intermediate feature distribution, aiming for improved generalization capabilities for unknown attacks. It includes the distance from the sample to the center of the sample’s class. $c_{y_i}$ denotes the $y_i$-th class center of features from face-cropped images. The center loss is calculated for each class (\textit{i.e.,} live and attack) regardless of the domain.
\begin{equation}\label{eq:centerloss}
\begin{aligned}
\mathcal{L}_{center} = \mathop{\mathbb{E}}\limits_{(\bar{x},y)\sim (\bar{\mathbf{X}},\mathbf{Y})}\left[\frac{1}{2}\left\|F(C(\bar{x})) - c_{y}  \right\|_{2}^{2}\right].
\end{aligned}
\end{equation}

\noindent\textbf{Total Loss:}
Finally, our total loss for PMN can be written using Equations.~\eqref{eq:ce_patch}, \eqref{eq:ce_whole}, and \eqref{eq:centerloss}, as follows:
\\
\begin{equation}\label{eq:loss_final}
\mathcal{L}_{PMN} = \mathcal{L}_{clip}  + \mathcal{L}_{face} + \mathcal{L}_{patch} + \alpha \mathcal{L}_{center} + \beta l_2 ,
\end{equation}
where $l_2$ is the $l_2$-regularization, and $\alpha$ and $\beta$ are the user-defined hyperparameters.
 
\section{Experiments\label{sec.4}}
\subsection{Experimental Setup}
\noindent\textbf{Dataset:} We evaluate the five benchmark datasets including CASIA-FASD~\cite{zhang2012face} (denoted as C), Idiap REPLAY-ATTACK~\cite{Chingovska_BIOSIG-2012} (denoted as I), OULU-NPU~\cite{OULU_NPU_2017} (denoted as O), MSU-MFSD~\cite{Diwen2014} (denoted as M), Rose Youtu~\cite{roseyoutube}\rev{, CASIA-SURF CeFA~\cite{cefa}, CASIA-SURF~\cite{surf}, and WMCA~\cite{wmca}}.
For all the tables in the rest of the manuscript, the datasets including OULU \cite{OULU_NPU_2017}, MSU-MFSD \cite{Diwen2014}, Idiap Replay-attack \cite{Chingovska_BIOSIG-2012}, and CASI-FASD \cite{zhang2012face} are abbreviated as $\{O, M, I, C\}$ respectively. \rev{Similarly, CASIA-SURF CeFA, CASIA-SURF, and WMCA are abbreviated as $\{C, S, W\}$.}

\noindent\textbf{Metric:} To evaluate performance, we use various metrics commonly employed in FAS, i.e., Attack Presentation Classification Error Rate (APCER), Bona Fide Presentation Classification Error Rate (BPCER), and Average Classification Error Rate (ACER) to ensure fairness in comparison. For cross-domain testing, we adopt ACER and Area Under Curve (AUC) as evaluation metrics.
\begin{table*}[t!]
\caption{
\label{tab:fourdataset}\rev{\textbf{Best-epoch comparison on the DG-FAS benchmark}.
ACER and AUC are reported under four DG protocols (O, M, I, C denote OULU, MSU-MFSD, Replay-Attack, and CASIA-FASD).
Bold and underlined values indicate the best and second-best performance, respectively. * denotes the use of CelebA-Spoof.}} 
\centering
\resizebox{1\linewidth}{!}{
\begin{NiceTabular}{@{}c|cc|cc|cc|cc|cc@{}}
\hline
\multirow{2}{*}{Method} & \multicolumn{2}{c|}{OCI$\rightarrow$M} & \multicolumn{2}{c|}{OMI$\rightarrow$C} & \multicolumn{2}{c|}{OCM$\rightarrow$I} & \multicolumn{2}{c|}{ICM$\rightarrow$O} & \multicolumn{2}{c}{Average} \\ 
\cline{2-11} 
 & ACER (\%) & AUC (\%) & ACER (\%) & AUC (\%) & ACER (\%) & AUC (\%) & ACER (\%) & AUC (\%) & ACER (\%) & AUC (\%) \\ 
\hline
MA-Net \cite{ma-net} & 20.80 & - & 25.60 & - & 24.70 & - & 26.30 & - & 24.35 & - \\
SSAN-R \cite{styleassemble} & 6.67 & 98.75 & 10.00 & 96.67 & 8.88 & 96.79 & 13.72 & 93.63 & 9.82 & 96.46 \\
AFD \cite{huang2023face} & 12.92 & 93.29 & 17.78 & 88.10 & 18.75 & 91.92 & 15.90 & 90.54 & 16.34 & 90.96 \\
PatchNet \cite{patchnet} & 7.10 & 98.46 & 11.33 & 94.58 & 13.40 & 95.67 & 11.82 & 95.07 & 10.90 & 95.95 \\
DFDN \cite{ma2024dual} &  5.20 & 98.39 & 8.00 & 97.45 & 7.71 & 95.56 & 11.01 & 95.22 & 7.98 & 96.66 \\
SAFAS \cite{sun2023rethinking}   & 5.95 & 96.55 & 8.78 & 95.37 & 6.58 & 97.54 & 10.00 & 96.23 & 7.83 & 96.42 \\

CA-FAS~\cite{long2025confidence} & 7.14 & 97.42 & 11.68 & 94.55 & 13.86 & 93.67 & 11.67 & 94.53 & 11.09 & 95.04 \\
ViT\&FA\&CS~\cite{cai2025towards} & 4.62 & 98.92 & 7.28 & 97.02 & 10.89 & 97.05 & 6.77 & 98.25 & 7.39 & 97.81\\
AG-FAS~\cite{long2024generalized} & 5.71 & 98.03 & 5.44 & 98.55 & 6.71 & 98.23 & 9.43 & 96.62 & 6.82 & 97.86 \\
CA-MoEiT~\cite{liu2024moeit} & 2.88 & 98.76& 7.89 & 97.70 & 6.18 & 98.94 & 9.72 & 96.22 & 6.67 & 97.91\\
FSFM ViT-B~\cite{wang2025fsfm} & 3.78 & 99.15 & 3.16 & 99.41 & 4.63 & 99.03 & 7.68 & 97.11 & 4.81 & 98.68 \\
CCPE~\cite{guo2025domain} & 3.10 & 99.21 & 1.33 & 99.36 & 6.08 & 94.36 & 5.57 & 98.49 & 4.02 & 97.86 \\


CFPL~\cite{liu2024cfpl}& 3.09&  \bf{99.45}& 2.56&  99.10& 5.43&  98.41& 3.33&  99.05& 3.60&99.00\\

\hline
FLIP-MCL*~\cite{srivatsan2023flip} & 5.00 & 98.35 & \bf{0.54} & \bf{99.98} & \underline{4.25} & \underline{99.07} & 3.70 & \underline{99.28}& 3.37 & \underline{99.17} \\
CFPL*~\cite{liu2024cfpl}& \bf{1.43}& 99.28&  2.56&  99.10& 5.43& 98.41& \bf{2.50}&  \bf{99.42}& \underline{2.98}&99.05\\
\hline
 PCGAN only(Ours) & 7.58 & 96.94 & 2.22 & 99.72 & 4.88 & 99.22 & 4.69 & 99.10 & 4.84 & 98.75\\
Ours     & \underline{2.50}& \underline{99.35}& \underline{2.04} & \underline{99.32} & \bf{3.33} & \bf{99.11} & \underline{3.29}& 99.08 & \bf{2.79} & \bf{99.22} \\
\bottomrule
\end{NiceTabular}
}
\vspace{-3mm}
\end{table*}

\noindent\textbf{Detailed Experimental Configuration:} Our experiments are conducted on an Nvidia RTX A6000 GPU with a batch size of 1, using the Adam optimizer for 4000 iterations. The initial learning rate is 1e-6 with betas {0.9, 0.999}. For face region cropping, we use dataset-provided face location information or MTCNN~\cite{zhang2016joint}, and apply a padding value of 0.6. The hyperparameters $\alpha$ and $\beta$ are set to 0.2 and 1e-6, respectively. Testing is performed using face-cropped images only.

\subsection{Quantitative Comparison}
This section presents the quantitative results of our method compared to various approaches across different cross-domain settings. First, we evaluate models with multiple training source domains in Tab~\ref{tab:fourdataset}. 
\rev{Second, following~\cite{sun2023rethinking}, we report the average performance over the last 10 epochs and evaluate learning stability by measuring the gap between the best epoch and the averaged results, as shown in parentheses in Tab~\ref{tab:last10}. Third, we extend our evaluation to large-scale cross-domain and cross-ethnicity benchmarks, including CASIA-SURF CeFA, CASIA-SURF, and WMCA (CSW benchmarks). 
Compared to conventional DG-FAS protocols based on small-scale RGB datasets, these benchmarks introduce substantially higher intra-class variation and domain shifts. 
The results in Tab.~\ref{tab:csw} show that our method maintains strong performance under these challenging settings, confirming its scalability and robustness in more realistic scenarios.}

\subsubsection{Cross-Domain Testing}
Tab.~\ref{tab:fourdataset} reports ACER and AUC results of various methods under four cross-domain settings. Our method achieves the lowest ACER of 3.33\% on OCM→I, and the second-best performance on the remaining three cases, where the best results are obtained by methods using CelebA-Spoof as extra data. Notably, among methods without CelebA-Spoof, our approach achieves the best ACER across all four cases. Moreover, considering both average ACER and AUC, our method outperforms all compared approaches, demonstrating state-of-the-art performance among CLIP-based methods.

\begin{table*}[t]
\caption{\label{tab:last10}
\rev{\textbf{Domain generalization results on the FAS benchmark}.
ACER and AUC are averaged over the last 10 epochs. Parentheses indicate ACER differences from Tab.~\ref{tab:fourdataset} as a measure of learning stability.
The same four protocols as Tab.~\ref{tab:fourdataset} are used.}}
\centering
\resizebox{1\linewidth}{!}{
\begin{tabular}{@{}c|cc|cc|cc|cc|cc@{}}
\hline
\multirow{2}{*}{Method} & \multicolumn{2}{c|}{OCI$\rightarrow$M} & \multicolumn{2}{c|}{OMI$\rightarrow$C} & \multicolumn{2}{c|}{OCM$\rightarrow$I} & \multicolumn{2}{c|}{ICM$\rightarrow$O} & \multicolumn{2}{c}{Average} \\ 
\cline{2-11} 
 & ACER (\%) & AUC (\%) & ACER (\%) & AUC (\%) & ACER (\%) & AUC (\%) & ACER (\%) & AUC (\%) & ACER (\%) & AUC (\%) \\ 
\hline
SSAN-R \cite{styleassemble}  & 21.79(15.12) & 84.06 & 26.44(16.44) & 78.44 & 35.39(26.51) & 70.13 & 25.72(12.00) & 79.37 & 27.34(17.52) & 78.00 \\
PatchNet \cite{patchnet} & 25.92(18.82) & 83.43 & 36.26(24.93) & 71.38 & 29.75(16.35) & 80.53 & 23.49(11.67) & 84.62 & 28.86(17.96) & 79.99 \\
SAFAS~\cite{sun2023rethinking}    & 14.36(8.41) & 92.06 & 19.40(10.62) & 88.69 & 11.48(4.90) & 95.74 & 11.29(1.29) & 95.23 & 14.13(6.26) & 92.93 \\
FLIP-MCL~\cite{srivatsan2023flip} & 15.00(10.00) & 93.37 & {\bf{3.22}}(2.68) & \bf{99.44} & 10.25(6.00) & 96.56 & 5.75(2.05) & \bf{98.48} & 8.56(5.19) & 96.96 \\
\hline
Ours     & {\bf{8.21}}(5.71) & \bf{96.95} & 5.41(3.37) & 98.76 & {{\bf9.55}}(6.22) & \bf{96.72} & {\bf{5.51}}(2.22) & 98.23 & {\bf{7.17}}(4.38) & \bf{97.67} \\
\hline
\end{tabular}}
\vspace{-5mm}
\end{table*}

\subsubsection{Cross-Domain Testing of the average results}
In cross-domain FAS, performance can vary significantly across epochs since the target domain is unseen during training. Following \cite{sun2023rethinking}, we report the average ACER and AUC over the last 10 epochs to assess generalization. Tab.~\ref{tab:last10} compares our method with others under this protocol.
For average ACER, our method achieves state-of-the-art performance in all cases except OMI→C. Notably, it records 8.21\% on OCI→M, outperforming the second-best result by a large margin, and achieves the lowest error rates on OCM→I (9.55\%) and ICM→O (5.51\%). On OMI→C, our method attains the second-best result (5.41\%), following FLIP~\cite{srivatsan2023flip}. Across all protocols, only our method consistently achieves ACER below 10\%.
For average AUC, our method attains the best performance on OMI→C and ICM→O. Overall, considering both ACER and AUC averages, our approach significantly outperforms competing methods, demonstrating strong and stable generalization.

\begin{table*}[t]
\caption{\label{tab:csw}\rev{\textbf{Results on CSW benchmarks}.
Evaluation metrics are ACER and AUC under cross-domain settings on CASIA-Surf-CeFA (C), CASIA-Surf (S), and WMCA (W), where bold and underlined values indicate the best and second-best performance, respectively.}}
\centering
\resizebox{0.83\linewidth}{!}{
\begin{NiceTabular}{@{}c|cc|cc|cc|cc@{}}
\hline
\multirow{2}{*}{Method} & \multicolumn{2}{c|}{CS$\rightarrow$W} & \multicolumn{2}{c|}{SW$\rightarrow$C} & \multicolumn{2}{c|}{CW$\rightarrow$S} & \multicolumn{2}{c}{Average} \\ 
\cline{2-9} 
& ACER (\%) & AUC (\%) & ACER (\%) & AUC (\%) & ACER (\%) & AUC (\%) & ACER (\%) & AUC (\%) \\ 
\hline
ViT  & 21.04 & 89.12 & 17.12 & 89.05 & 17.16 & 90.25 & 18.44 & 89.47 \\
CLIP-V~\cite{radford2021learning} & 20.00 & 87.72 & 17.67 & 89.67 & 8.32 & 97.23 & 15.33 & 91.54\\
CLIP~\cite{radford2021learning} & 17.05 & 89.37 & 15.22 & 91.99 & 9.34 & 96.62 & 13.87 & 92.66 \\
CA-MoEiT~\cite{liu2024moeit} & 16.67 & 91.02 & 16.42 & 93.17 & 12.57 & 93.76 & 15.22 & 92.59 \\
CoOp~\cite{zhou2022learning} & \textbf{9.52} & 90.49 & 18.30 & 87.47 & 11.37 & 95.46 & 13.06 & 91.14\\
CFPL~\cite{guo2025domain} & 9.57 & \textbf{94.25} & 14.89 & 91.56 & 8.16 & 96.78 & 10.87 & 94.20 \\
\hline
 Ours & 12.58 & 94.10 & \textbf{11.88} & \textbf{94.96} & \textbf{7.89} & \textbf{97.67} & \textbf{10.78} & \textbf{95.58} \\
\hline
\end{NiceTabular}}
    \vspace{-3mm}
\end{table*}

\begin{table}[t!]
    \centering
    \caption{\label{tab:ablationstudy}
    For the ablation study, we report results on a domain-generalization FAS benchmark using ACER. A checkmark indicates the inclusion of each component in PMN and PCGAN, where CL, PL, and Syn denote Center Loss, Patch Loss, and synthesized images, respectively.
    }
    \resizebox{0.7\linewidth}{!}{
    \begin{tabular}{c|c|c|cccc|c}
        \hline
        \multicolumn{8}{c}{(a) Protocol 1: Best epochs}\\   
        \hline
        \multicolumn{2}{c|}{\textbf{\rev{PMN}}} & \textbf{\rev{PCGAN}} & \multirow{2}{*}{\textbf{OCI$\rightarrow$M}} & \multirow{2}{*}{\textbf{OMI$\rightarrow$C}} & \multirow{2}{*}{\textbf{OCM$\rightarrow$I}} & \multirow{2}{*}{\textbf{ICM$\rightarrow$O}} & \multirow{2}{*}{\textbf{Avg.}} \\ 
        \textbf{CL} & \textbf{\rev{PL}} & \textbf{Syn} & & & & &  \\ 
        \hline
        & & & \rev{10.42} & \rev{4.44} & \rev{7.75}& \rev{6.11} & \rev{7.18} \\ 
        & \rev{\checkmark} & & 10.00 & 2.04 & 6.67 & 3.38 & 5.52\\
        \checkmark& & & \rev{7.50} & \rev{2.22} & \rev{5.25}  & \rev{5.14} & \rev{5.03}\\ 
        \checkmark & \rev{\checkmark} & & 5.00 & 3.33 & 6.67 & 4.81 & 4.95\\
        & & \checkmark & \rev{7.58} & \rev{2.22} & \rev{4.88} & \rev{4.69} & \rev{4.84} \\ 
        & \rev{\checkmark} & \checkmark & 7.92 & 2.22 & 6.83 & \textbf{2.59} & 4.89\\
        \checkmark& & \checkmark & \rev{5.42}& \rev{2.41} & \rev{6.12} & \rev{4.48} & \rev{4.61}\\ 
        \checkmark & \rev{\checkmark} & \checkmark & \textbf{2.50} & \textbf{2.04} & \textbf{3.33} & 3.29 & \textbf{2.79} \\
        \hline
    \end{tabular}}
    \vspace{-3mm}
\end{table}

\subsubsection{Learning Stability:}
Tab~\ref{tab:fourdataset} shows the best results, and Tab~\ref{tab:last10} displays the average results for the last 10 epochs. 
In the FAS task, the model is challenging to converge, so the average of the last 10 epochs is considered an approximate convergence value. The difference in ACER values between Tab~\ref{tab:fourdataset} and Tab~\ref{tab:last10}, as indicated in parentheses on Tab~\ref{tab:last10}, represents how well the model converges close to the best results.
For overall methods, while learning stability is not good in the case of OMI$\rightarrow$C and OCM$\rightarrow$I, learning stability is good in the OCI$\rightarrow$M and ICM$\rightarrow$O.
The averages of OCI$\rightarrow$M and ICM$\rightarrow$O datasets are more stable than OMI$\rightarrow$C and OCM$\rightarrow$I datasets for learning.
Our methods achieve the best stability result for one case and the second-best result for two cases. 
In terms of the average, our method attains the highest learning stability score.

\subsection{\rev{Large-Scale Cross-Domain Evaluation on CSW Benchmarks}}
\rev{
Tab~\ref{tab:csw} presents the ACER and AUC performance of various methods on the CSW large-scale benchmarks under three cross-domain settings: CS$\rightarrow$W, SW$\rightarrow$C, and CW$\rightarrow$S. 
Compared with conventional small-scale RGB datasets, these benchmarks introduce larger subject diversity and more severe domain shifts, making them substantially more challenging.
In terms of ACER, our method achieves the best average performance across the three protocols, with an average ACER of 10.78\%, outperforming all compared approaches. 
Notably, our method obtains the lowest error rate of 7.89\% in the CW$\rightarrow$S setting and demonstrates competitive performance in the other two protocols. 
Regarding AUC, our method consistently achieves the highest average score 95.58\%, indicating more reliable discrimination capability under large-scale cross-domain conditions.
These results demonstrate that the proposed PCGAN and PMN framework generalizes effectively beyond traditional DG-FAS benchmarks and maintains strong robustness under more realistic and diverse evaluation scenarios.
}

\subsection{Quantitative Analysis}
\subsubsection{Effectiveness of components}
\rev{
This paragraph analyzes the contributions of three components in our framework: Center Loss (CL), Patch-based Learning (PL) in PMN, and synthesized images (Syn) generated by PCGAN. As shown in Tab.~\ref{tab:ablationstudy}, the baseline without these components performs worst, confirming the need for additional supervision.
Introducing PL consistently improves performance, demonstrating the effectiveness of patch-level supervision for localized spoof artifacts. Incorporating Syn further enhances cross-domain generalization, especially in challenging protocols such as OCM→I and ICM→O, while CL stabilizes feature distributions and improves robustness.
The full model combining CL, PL, and Syn achieves the best overall performance, highlighting strong complementarity among the components: PL enables localized artifact reasoning, Syn increases domain diversity, and CL promotes feature compactness, resulting in robust and stable generalization.
}

\begin{table}[t]
    \caption{\label{tab:disen}\rev{Ablation study on synthetic-live and synthetic-spoof samples generated by PCGAN.
Spoof→Live replaces real live data with artifact-removed spoof images, while Live→Spoof replaces real spoof data with artifact-injected live images.
ACER (\%) is reported under four cross-domain protocols.}}
    \centering
    \resizebox{0.65\linewidth}{!}{
    \begin{NiceTabular}{cc|cccc|c}
        \hline
        Spoof$\rightarrow$Live & Live$\rightarrow$Spoof & OCI$\rightarrow$M & OMI$\rightarrow$C & OCM$\rightarrow$I & ICM$\rightarrow$O & Average \\
        \hline
         & & 5.00 & 3.33 & 6.67 & 4.81 & 4.95 \\
        \checkmark & & 5.42 & 4.44 & 8.75 & 4.17 & 5.70\\
        & \checkmark & 7.50 & 6.67 & 10.00 & 3.82 & 7.00 \\
        \checkmark & \checkmark & 7.50 & 3.33 & 11.25 & 3.02 & 6.28 \\
        \hline
    \end{NiceTabular}}
    \vspace{-3mm}
\end{table}

\rev{
\subsubsection{Effectiveness of PCGAN-Generated Synthetic Samples}
To evaluate PCGAN under a controlled setting, we conduct an ablation study where the detector is trained without any real live images, using only original spoof images and PCGAN-generated synthetic samples. This design isolates the effect of artifact removal and injection from real live data.
As shown in Tab.~\ref{tab:disen}, the baseline achieves an average ACER of 4.95\%. Replacing real live images with synthetic-live samples generated via Spoof→Live yields comparable performance (5.70\% ACER), despite the absence of real live data, indicating that PCGAN effectively removes spoof artifacts while preserving semantic facial content.
In contrast, training with synthetic-spoof samples from Live→Spoof results in a larger degradation (7.00\% ACER), suggesting that artifact injection alone is insufficient to model real spoof diversity. Using both synthetic-live and synthetic-spoof samples improves performance (6.28\% ACER) but remains inferior to the Spoof→Live setting, highlighting the asymmetric difficulty between artifact removal and injection.
Overall, these results demonstrate that PCGAN learns controllable artifact-specific representations, with Spoof→Live enabling effective cross-domain training even without real live data.
}

\begin{table}[t]
    \caption{\label{tab:able_oulusyn}ACER (\%) performance on different datasets. We train the model using either the original images (Original OULU) or synthesized images (Syn OULU).} 
    \centering
    \resizebox{0.5\linewidth}{!}{
    \begin{NiceTabular}{cc|cccc|c}
        \hline
        \multicolumn{7}{c}{(a)Protocol 1: Best epochs} \\
        \hline
        Origin & Syn & OULU & MSU & Idiap & CASIA & Average \\
        \hline
        \checkmark & & \bf{0.00} & \bf{5.00} & 7.83 & 4.63 & 4.37 \\
        & \checkmark & 8.33 & 25.00 & 16.33 & 7.96 & 14.41 \\
        \checkmark & \checkmark & \bf{0.00} & 7.50 & \bf{5.17} & \bf{4.44} & \bf{4.28} \\
        \hline
    \end{NiceTabular}}
    \vspace{-3mm}
\end{table}

\begin{table}[t]
    \caption{
    Experimental results show the impact of changing the random cropping size. The metrics used in this experiment are the ACER.\label{tab:appendixc}
    }
    \centering
    \resizebox{0.5\linewidth}{!}{
        \begin{tabular}{@{}c|cccc|c@{}}
        
        \midrule
        \textbf{Range}            & \textbf{OCI$\rightarrow$M} & \textbf{OMI$\rightarrow$C} & \textbf{OCM$\rightarrow$I} & \textbf{ICM$\rightarrow$O} & \textbf{Avg.} \\ 
        \midrule
        0.2-1.0 & \bf{2.50} & 2.04 & \bf{3.33} & 3.29 & \bf{2.79} \\
        0.2-0.2 & 5.00 & \bf{1.11} & 8.50 & 3.38 & 4.50 \\
        0.4-0.4 & 5.42 & 2.22 & 11.67 & \bf{2.22} & 5.38 \\
        0.6-0.6 & 7.08 & 2.22 & 10.00 & 3.38 & 5.67 \\
        0.8-0.8 & 5.00 & \bf{1.11} & 10.00 & 3.33 & 4.86 \\
        
        \bottomrule
        \end{tabular}
    } \vspace{-3mm}
\end{table}

\subsubsection{Generalization through PCGAN-based Augmentation:}
We set up scenarios in which we train the model on the original image, synthesized image, and both, respectively, for the OULU dataset~\cite{OULU_NPU_2017}. Subsequently, we validate each model on the test datasets for all datasets.
Tab~\ref{tab:able_oulusyn} shows the results of the above scenario.
Given training on the original dataset as the baseline, training on the synthesized dataset decreases performance compared to the baseline, while training on both improves the model's performance. Synthesized datasets show good synergy with original datasets, although they are not useful on their own. Therefore, images synthesized from PCGAN are effective enough as augmented datasets.

\subsubsection{Effectiveness of cropping size}
In this section, we analyze the effect of face cropping size on patch-based learning. As shown in Tab.~\ref{tab:appendixc}, we compare random cropping with a scale range of 0.2–1.0 against fixed cropping scales (0.2, 0.4, 0.6, 0.8). Fixed cropping achieves comparable or better performance on OMI→C and ICM→O, but degrades on OCI→M and OCM→I, with particularly severe drops on OCM→I. In contrast, random cropping shows stable performance across all datasets. Although specific fixed scales perform best in certain environments, these results suggest that optimal crop sizes are dataset-dependent, motivating the use of random cropping to cover diverse scales.

\begin{figure*}[t!] 
    \centering
    \includegraphics[width=1\linewidth]{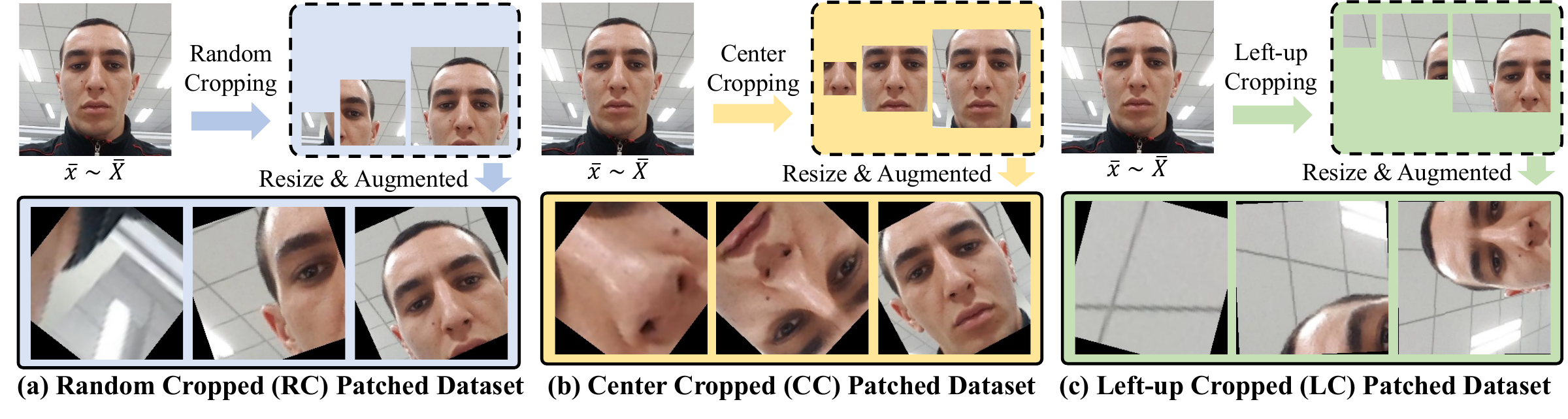}
    \vspace{-15mm}
    \caption{Examples displayed each cropping approach.}
    \label{fig:crop}
    \vspace{-3mm}
\end{figure*}

\begin{table}[t!]
\centering
\caption{
Experimental results on the domain-generalization for FAS benchmark. The metrics used in this experiment are the ACER. RC, CC, and LC mean Random Cropping, Center Cropping, and Left Up Cropping, respectively.
}
\label{tab:appendixa}
\resizebox{0.5\linewidth}{!}{
\begin{tabular}{@{}c|cccc|c@{}}
\toprule
\multicolumn{6}{c}{(a) Protocol 1: Best epochs}\\

\midrule
\textbf{Method}            & \textbf{OCI$\rightarrow$M} & \textbf{OMI$\rightarrow$C} & \textbf{OCM$\rightarrow$I} & \textbf{ICM$\rightarrow$O} & \textbf{Avg.} \\ 

\midrule
RC & \bf{2.50} & 2.04 & \bf{3.33} & 3.29 & \bf{2.79} \\
CC & 7.50 & \bf{1.11} & 8.33 & 4.77 & 5.43 \\
LC & 4.17 & 2.22 & 10.00 & \bf{2.59} & 4.74 \\


\bottomrule
\end{tabular}
} 
\vspace{-3mm}
\end{table}

\subsubsection{Various Cropping Approaches}
In this section, we analyze the impact of different cropping strategies on our method. To learn localized facial features (e.g., eyes and nose), we construct a patched dataset using various cropping approaches and conduct comparative experiments.

\noindent\textbf{Random Cropping (RC)}
RC randomly selects cropping regions (Fig.~\ref{fig:crop}-(a)) with scales ranging from 0.2 to 1.0 of the image size, enabling the model to learn multi-scale localized features from diverse spatial locations.

\noindent\textbf{Center Cropping (CC)}
CC selects cropping regions centered in the image (Fig.~\ref{fig:crop}-(b)) with scales ranging from 0.2 to 1.0, encouraging the model to learn multi-scale localized features around central facial regions.

\noindent\textbf{Left-up Cropping (LC)}
LC selects cropping regions from the upper-left area of the image (Fig.~\ref{fig:crop}-(c)) with scales ranging from 0.2 to 1.0, promoting the learning of multi-scale localized features biased toward the upper-left facial region.

\noindent\textbf{Quantitative Results}
Tab.~\ref{tab:appendixa} reports ACER results for different cropping strategies under four cross-domain settings. RC achieves the best performance on OCI→M and OCM→I, and the second-best on OMI→C and ICM→O, while CC and LC perform best on OMI→C and ICM→O, respectively. Overall, RC shows the most consistent performance across environments, indicating stronger generalization.

\begin{figure}[t] 
    \centering
    \includegraphics[width=0.65\linewidth]{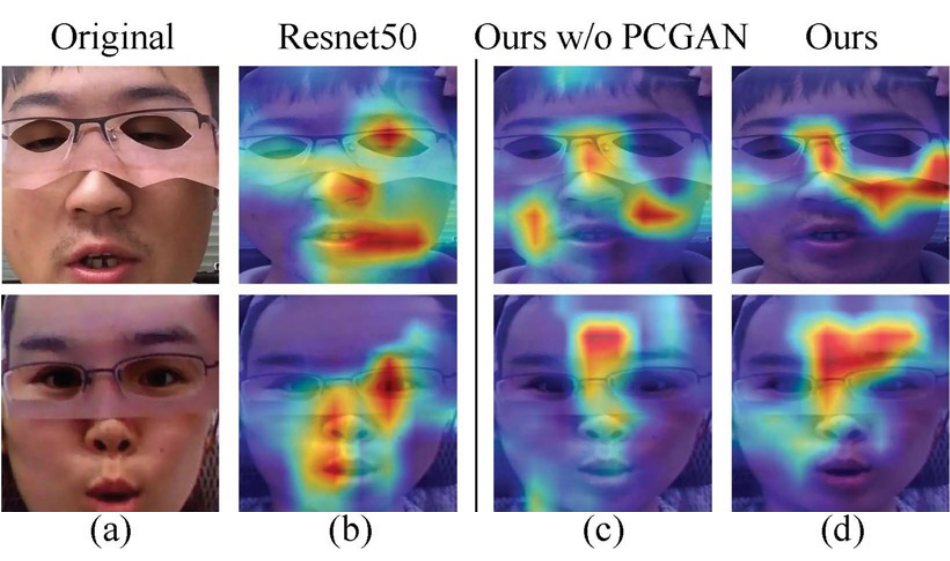}
    \vspace{-7mm}
    \caption{Grad-CAM visualizations under the MOI→C protocol.
Activation maps for the attack class on the ROSE-Youtu dataset are shown to evaluate robustness to unseen partial attacks. (a) Attack image. (b) ResNet50 without PMN and PCGAN. (c) Our model with PMN. (d) Our model with both PMN and PCGAN.}
    \vspace{-3mm}
    \label{fig:gradcam}
\end{figure}

\subsection{Qualitative Analysis}
\noindent\textbf{Grad-CAM Visualization:}
Fig~\ref{fig:gradcam} illustrates the grad-cam visualization to show the effect of the patch-based learning and synthesized data.
Ours w/ patch, Ours w/ patch$\&$syn, and ResNet w/o patch denote the proposed method employing patch-based learning, patch-based learning incorporated with synthetic samples, and baseline.
For case (a-b), ResNet w/o patch falsely classifies the spoof images, partial attack case, to live. Furthermore, even for the correct classification, as shown in (c-d), ResNet w/o patch highlights the region unnecessary for the decision. 
Conversely, when applying our proposed method, without using the synthesized samples for training, the grad-cam directs semantically plausible regions to classify the spoof image. 
Also, the grad-cam results in row (d) show that the synthetic samples help the network focus more on spoofed regions.
The grad-cam visualizations provide substantial clues why our proposed method performs robust classification for complex presentation cases such as partial attacks.

\begin{figure}[t]
    \centering
    \includegraphics[width=0.65\linewidth]{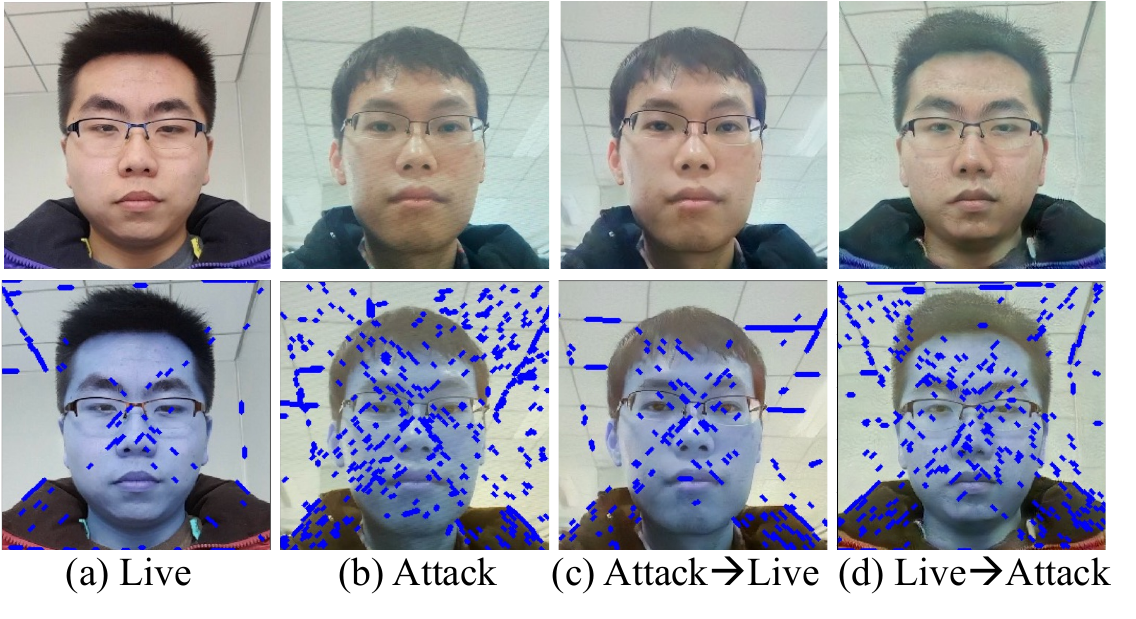}
    \vspace{-7mm}
    \caption{Cross class results with Pattern Conversion GANs. The images show the artifact pattern detection results by Canny detection and Hough Transform. (c) shows the artifact-pattern removal results from (b). (d) inserts artifact patterns from (a) into (b).}
    \label{fig:sae_visual}
    \vspace{-3mm}
\end{figure}

\begin{figure}[t!]
    \centering
    \includegraphics[width=0.65\linewidth]{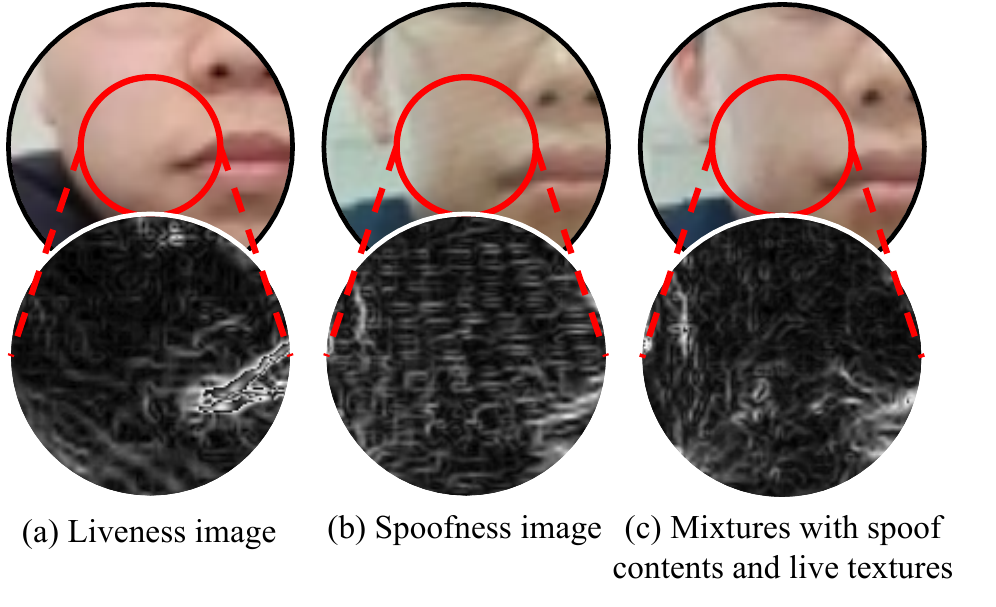}
    \vspace{-7mm}
    \caption{\textbf{Distinct visual artifacts extracted via Sobel filtering.}
(a)–(c) are taken from Fig.~\ref{fig:framework}.}
    \label{fig:difference}
    \vspace{-3mm}
\end{figure}

\noindent\textbf{Conversion results from Artifact Patterns:}
This chapter visualizes PCGANs disentanglement of artifact patterns from spoof images. Fig~\ref{fig:sae_visual} shows the result of adding the artifact patterns of the attack image to the live image or separating and removing the artifact patterns from the attack image. artifact patterns in the example are moiré artifacts caused by the display and have linear characteristics. Therefore, after detecting edge components through Canny Detection, linear components can be visualized through Hough Transform. As shown in the results, it can be seen that the artifacts detected in the spoofed image are reduced, and the artifacts are injected into the real image.

Fig.~\ref{fig:difference} demonstrates the example of spoof textures exhibiting recurrent simple patterns originating from the Moiré effect. To emphasize the patterns on the second row, we obtained the gradient magnitude map using 3$\times$3 Sobel filters in the x and y directions. In Fig.~\ref{fig:difference}, (a) does not reveal the recurrent pattern, (b) clearly shows the pattern, and (c) demonstrates that the pattern has been removed by PCGANs.

\begin{table}[t]
    \caption{
    \rev{\textbf{Computational cost comparison in terms of parameters and FLOPs}.
The generator is used only during training, so inference cost reflects the detector alone.\label{tab:cost}}
    }
    \centering
    \resizebox{0.9\linewidth}{!}{
        \begin{NiceTabular}{@{}c|ccc|cc|cc|ccc@{}}
        \hline
        \multirow{3}{*}{Method} & \multicolumn{3}{c|}{\multirow{2}{*}{Generator}} & \multicolumn{4}{c|}{Detector} & \multicolumn{2}{c}{\multirow{3}{*}{Inference}} \\
        &  &  & & \multicolumn{2}{c|}{Image Encoder} & \multicolumn{2}{c|}{Text Encoder} & \\
        & Type & Parameters & Flops & Parameters & Flops & Parameters & Flops & Parameters & Flops & Inference Time\\
        \hline
        ViT~\cite{li2022exploring} & None & - & - & 86.19M & 17.58G & - & - &  86.19M & 17.58G & 0.007\\
        FLIP-IT~\cite{srivatsan2023flip} & None & - & - & 86.19M & 17.58G & 63.11M & 35.81M &  86.19M & 17.58G & 0.0010\\
        FLIP-MCL~\cite{srivatsan2023flip} & None & - & - & 86.19M & 52.74G & 83.05M & 35.86M &  86.19M & 17.58G & 0.0010\\
        Ca-MoEiT~\cite{liu2024moeit} & None & - & - & 86.19M & 17.58G & - & - &  86.19M & 17.58G & - \\
        AG-FAS~\cite{long2024generalized} & Diffusion & 1.07B & 16.94T & 86.19M & 17.58G & - & - &  86.19M & 17.58G & -\\
        Ours & GAN & 109.03M & 95G & 86.19M & 35.16G & 63.11M & 35.81M &  86.19M & 17.58G & 0.0010 \\
        \hline
        \end{NiceTabular}
    }\vspace{-3mm}
\end{table}
\rev{
\subsection{Computational Cost Analysis}
Tab.~\ref{tab:cost} compares the computational complexity of our method with recent state-of-the-art approaches. We report parameters and FLOPs for the generator (if applicable), detector, and inference stage. Although our framework includes a GAN-based generator (109.03M parameters, 95G FLOPs), it is used only during training and excluded at inference. Thus, the runtime cost equals that of the CLIP-based detector backbone (86.19M parameters, 17.58G FLOPs). Compared to diffusion-based methods such as AG-FAS, which require substantially higher generation cost, our approach significantly reduces training-time complexity while maintaining identical inference cost.
}\vspace{-10mm}

\section{Conclusion\label{sec.5}}
\rev{
In this paper, we proposed a domain-generalizable face anti-spoofing framework that integrates a Pattern Conversion GAN (PCGAN) with a Patch-based Multi-tasking Network (PMN). By disentangling spoofing artifacts from facial content and recombining them across samples, the proposed method alleviates the limited diversity of existing FAS datasets and improves robustness to unseen domains and partial attacks.
A key strength of our approach is its artifact-centric design. Unlike prior generative methods that focus on global appearance or domain styles, PCGAN explicitly models spoofing artifacts introduced by attack media and recapturing processes, enabling controllable artifact removal and injection. In addition, patch-based multi-task learning encourages the detector to jointly exploit global facial cues and localized evidence, leading to improved generalization in challenging cross-domain scenarios.
Despite these advantages, the proposed method has limitations. The diversity of synthesized samples is bounded by the artifacts present in the training data, and the generator is intended for data augmentation rather than large-scale or real-time synthesis. Moreover, although extensive experiments demonstrate strong generalization, further evaluation on emerging attack types and more unconstrained real-world scenarios remains an important direction for future work.
Nevertheless, this work provides a practical framework for improving domain generalization in face anti-spoofing. The proposed artifact disentanglement and patch-based learning strategies are readily extensible to other FAS backbones and potentially to broader biometric anti-spoofing tasks. Future work will explore richer artifact synthesis with advanced generative models, finer-grained spatial modeling, and extensions to video-based or multimodal FAS systems.
}\vspace{-3mm}


\section*{Acknowledgment}
This work was supported by Institute of Information \& communications Technology Planning \& Evaluation (IITP) grant funded by the Korea government (Ministry of Science and ICT) (RS-2021-II211341, Artificial Intelligence Graduate School Program(Chung-Ang University) and Graduate School of Metaverse Convergence support program (IITP-2023(2024)-RS-2024-00418847)).



\end{document}